\documentclass[runningheads,a4paper]{llncs} 

\usepackage[printonlyused,withpage]{acronym}
\usepackage{amsmath}
\usepackage{amssymb}
\usepackage{booktabs}
\usepackage{graphicx}
\usepackage{subfigure}
\usepackage{todonotes,varwidth}
\usepackage{paralist}
\usepackage{hyperref}
\usepackage{multirow}
\usepackage{comment}
\usepackage{float}

\newcommand{\W}{\mathbf{W}}
\newcommand{\X}{\mathbf{X}}
\newcommand{\Y}{\mathbf{Y}}

\usepackage{etoolbox,siunitx}
\robustify\bfseries

\newcommand{\x}{\mathbf{x}}

\newcommand{\y}{\mathbf{y}}
\newcommand{\w}{\mathbf{w}}

\newcommand{\alphab}{\boldsymbol{\alpha}}
\newcommand{\gammab}{\boldsymbol{\gamma}}
\newcommand{\betab}{\boldsymbol{\beta}}

\newcommand{\psib}{\boldsymbol{\psi}}
\newcommand{\Psib}{\boldsymbol{\Psi}}

\newcommand{\deltab}{\boldsymbol{\delta}}

\newcommand{\ie}{{\em i.e.~\/}}
\newcommand{\eg}{{\em e.g.~\/}}

\newcommand{\etc}{{\em etc.~\/}}

\newcommand{\cf}{{\em c.f.~\/}}

\renewcommand{\lim}{\operatornamewithlimits{lim}}

\acrodef{EPSRC}{Engineering and Physical Sciences Research Council}
\acrodef{SPHERE}{Sensor Platform for HEalthcare in Residential Environment}
\acrodef{IRC}{Inter\-disciplinary Research Collaboration}
\acrodef{CRF}{Conditional Random Field}
\acrodef{L-CRF}{Linear Chain \ac{CRF}}
\acrodef{IID}[i.i.d.]{Independent and Identically Distributed}
\acrodef{NLP}{Natural Language Processing}
\acrodef{SVM}{Support Vector Machine}
\acrodef{RBF}{Radial Basis Function}
\acrodef{RF}{Random Forest}
\acrodef{LR}{Logistic Regression}
\acrodef{POS}{Part of Speech}
\acrodef{LR}{Logistic Regression}

\acrodef{NN}{Neural Network}
\acrodef{WH}{Word Hyphenation}
\acrodef{ODC}{Occasionally Dishonest Casino}
\acrodef{AR}{Activity Recognition}

\acrodef{CASAS}[{\sc CASAS}]{Centre for Advanced Studies in Adaptive Systems}
\acrodef{PIR}{Passive Infra-Red}
\acrodef{ADL}{Activities of Daily Living}
\acrodef{IADL}{Instrumental \ac{ADL}}
\acrodef{MLE}{Maximum Likelihood Estimate}
\acrodef{RBF}{Radial Basis Function}
\acrodef{RKHS}{reproducing kernel Hilbert space}
\acrodef{HMM}{Hidden Markov Model}
\acrodef{ERM}{Empirical Risk Minimisation}
\acrodef{AC}{Auto-correlation}
\acrodef{SVD}{Singular Value Decomposition}
\acrodef{GP}{Gaussian Process}
\acrodef{GPC}{Gaussian Process Classifier}
\acrodef{BPM}{Bayes Point Machine}

\acrodef{MAE}{Mean Absolute Error}

\acrodef{TP}{True Positive}
\acrodef{TN}{True Negative}
\acrodef{FP}{False Positive}
\acrodef{FN}{False Negative}

\acrodef{LBFGS}[L-BFGS]{Limited-memory Broyden-Fletcher-Goldfarb-Shanno}

\acrodef{NLL}[$\mathcal{NLL}$]{Negative Log Likelihood}
\acrodef{linreg}{Linear Regression}
\acrodef{logreg}{Logistic Regression}
\acrodef{nest}{Nested Logistic Regression}
\acrodef{struct}{Structured Ordinal Classification}
\acrodef{AD}{Alzheimer's disease}
\acrodef{PD}{Parkinson's disease}
\acrodef{SH}{Smart Home}
\acrodef{MCI}{Mild Cognitive Impairment}
\acrodef{LIWC}{Linguistic Inquiry and Word Count}
\newcommand{\storm}{{\sc StORM}}
\newcommand{\ol}{{\sc OrdLog}}
\newcommand{\lr}{{\sc LogReg}}
\newcommand{\bn}{{\sc BinNest}}

\acrodef{CSC}{Cost Sensitive Classification}
\acrodef{OVA}[OvA]{One-Versus-All}
\acrodef{OVO}[OvO]{One-Versus-One}
\acrodef{OR}{Ordinal Regression}

\acrodef{OL}  [\ol]{Ordered Logit} 
\acrodef{LR}  [\lr]{Logistic Regression}
\acrodef{NEST}[\bn]{Nested Binary Ordinal Regression}
\acrodef{STORM} [\storm]{Structured Ordinal Regression Modeling}

\acrodef{CDD}{Critical Difference Diagram}
\acrodef{ECOC}{Error Correcting Output Codes}

\usepackage{balance}
\usepackage{mathtools}
\usepackage[noend]{algpseudocode}
\usepackage[toc,page]{appendix}

\newcommand{\linear}{{\sc Linear}}
\newcommand{\jagged}{{\sc Sine}}
\renewcommand{\circle}{{\sc Circle}}
\newcommand{\spiral}{{\sc Spiral}}

\newcommand{\casas}{{\sc CASAS}}
\newcommand{\dementiabank}{{\sc DementiaBank}}

\newcommand{\autompg}{{\sc AutoMPG}}
\newcommand{\diabetes}{{\sc Diabetes}}
\newcommand{\abalone}{{\sc Abalone}}
\newcommand{\bostonhousing}{{\sc BostonHousing}}
\newcommand{\machinecup}{{\sc MachineCup}}
\newcommand{\pyrimidines}{{\sc Pyrimidines}}
\newcommand{\stocksdomain}{{\sc StocksDomain}}
\newcommand{\triazines}{{\sc Triazines}}
\newcommand{\wisconsin}{{\sc Wisconsin}}

\title{Ordinal Regression as Structured Classification}

\author{Niall Twomey \and Rafael Poyiadzi \and Callum Mann \and Ra\'ul Santos-Rodr\'iguez}
\institute{
  University of Bristol, Bristol, United Kingdom \\
  \texttt{\{niall.twomey, rp13102, cm13558, enrsr\}@bristol.ac.uk} 
}

\begin{document}

\maketitle 

\title{Ordinal Regression as Structured Classification}



\begin{abstract}

This paper extends the class of \ac{OR} models with a structured interpretation of the problem by applying a novel treatment of encoded labels. The net effect of this is to transform the underlying problem from an \ac{OR} task to a (structured) classification task which we solve with conditional random fields, thereby achieving a coherent and probabilistic model in which all model parameters are jointly learnt. Importantly, we show that although we have cast \ac{OR} to classification, our method still fall within the class of decomposition methods in the \ac{OR} ontology. This is an important link since our experience is that many applications of machine learning to healthcare ignores completely the important nature of the label ordering, and hence these approaches should considered na\"{i}ve in this ontology. We also show that our model is flexible both in how it adapts to data manifolds and in terms of the operations that are available for practitioner to execute. Our empirical evaluation demonstrates that the proposed approach overwhelmingly produces superior and often statistically significant results over baseline approaches on forty popular \ac{OR} models, and demonstrate that the proposed model significantly out-performs baselines on synthetic and real datasets. Our implementation, together with scripts to reproduce the results of this work, will be available on a public GitHub repository. 

\end{abstract}

\section{Introduction}
\label{section:introduction}

\ac{OR} is the task of learning to classify data-points into one of many interval classes. 
It can be understood as lying in between the canonical problems of classification and regression, as it is a classification task where the classes follow a pre-defined order. Model learning in these domains therefore requires particular care and attention since many assumptions underpinning standard classifiers are unsuitable in \ac{OR} settings. 

Let us consider \ac{AD} as a motivating application for this work. When assessing the current state of \ac{AD}, healthcare professionals utilise one of several well-known assessment questionnaires (\cf \cite{katz1983assessing}). 
These questionnaires are designed to uncover the cognitive capacity of the persons and evaluate the risks of independent living. 
An emerging application area of machine learning has been to non-invasively predict questionnaire scores based on a person's behaviour and circadian patterns of \ac{ADL} and \ac{IADL} in a \ac{SH} \cite{dawadi2013automated,urwyler2017cognitive} or to assess the cognitive ability from conversation analysis. 
These are challenging problems to model, and there has been some success in these areas already. 

The standard machine learning approach is based on learning a mapping between samples and categories so that the probability of error is minimised. However, in the setting described here the categorisation of the scores into their groups is an ordinal operation (\eg `severe' diagnoses are more extreme than `moderate' and `mild'), and indeed classifying a person with `severe' \ac{AD} as `mild' is more costly than predicting `moderate'. 
Automated \ac{AD} assessment presents an opportunity to produce valuable healthcare technology that can benefit vulnerable persons and their families, but also to benefit clinicians via the unprecedented and objective view into the effect of \ac{AD} on routine and behaviour. 
Although in the authors' experience the vast majority 
of the experimental literature on ordinal medical domains ignores the ordinal nature of the data and recasts the problem into traditional binary or multiclass problems, with some notable exceptions \cite{doyle2014predicting}. 

In this work we introduce a structural interpretation of ordinal regression. The advantage of this interpretation is that significantly more flexibility is ascribed to the predictive model, and this flexibility permits the model to operate efficiently on linear and non-linear data manifolds, while the baseline methods considered were unable achieve this. Additionally, our structured interpretation captures contextual information that the other baselines cannot.

The aims and contributions of this paper are as follows: 
We strongly advocate the selection of ordinal techniques for ordinal problems and  a review of ordinal approaches in Section \ref{section:related_work}. 
We extend the class of ordinal regression models in this work with a new structural interpretation of the field (Section \ref{section:methods}), outline empirical experiments (Section \ref{section:experiments}) demonstrate its utility in our results (Section \ref{section:results}). 
We summarise and conclude in Sections \ref{section:discussion} and \ref{section:conclusion}.

\section{Ordinal Regression}
\label{section:related_work}
Within the published area of \ac{OR}, there are several methodologies that are well established. We describe these with strong reference to the `ordinal regression ontology' from \cite{gutierrez2016ordinal} and then introduce the proposed approach after. 


\subsection{Na\"{i}ve Models}
\label{section:related_work:naive}
Intuitively we can reduce an \ac{OR} task to either a classification or a regression problem. In the case of classification, we ignore the nature of the classes, and proceed with a model that uses nominal classification. This is considered a na\"{i}ve approach as the practitioner ignores prior knowledge (\eg of class ordering) that could otherwise be used to increase the accuracy and predictive power of the model. For the case of regression, one may map the classes on the real line, employ regression techniques, then map back to the original classes. Unless the practitioner has a well considered way of computing the forward and backward mappings, this approach appears na\"{i}ve.

A similar approach to the classification reduction, but more advanced, is that of \ac{CSC}. \ac{CSC} is a general treatment of models where the practitioner provides (potentially) unique penalties for each type of misclassification \cite{tu2010one}. This is usually accomplished through the use of a cost matrix during learning. \ac{CSC} can therefore be employed for \ac{OR} by devising a cost matrix that depends on the \textit{distance} between classes \cite{kotsiantis2004cost}. This would again be a sensible approach given that the practitioner has a good understanding of the \textit{distances} between classes, and a principled way of transforming them to suitable costs.

\subsection{Threshold Models}
Threshold models are another approach to \ac{OR}. We assume that there is a latent continuous random variable that gives rise to the observed discrete classes. With this formulation we can perform a reduction to a regression problem. As criticised earlier this would be a {na\"{i}ve} approach due to the lack of principled way of mapping from the real line to the given classes. 
Approaches under the Threshold Models category, aim to surpass this limitation by learning this map, or where to `cut' the real line from data, as opposed to assuming knowledge of it, \textit{a priori}.



\paragraph{Ordered Logit:} 
The classical ordered logit model \cite{mccullagh1980regression} is a simple model that assumes a real-valued latent variable ($y^*$) is defined by

\begin{align}
    y^* = \w^\top \x + \epsilon
\end{align}

\noindent where $\x \in \mathbb{R}^D$ is a data point, $D$ is the dimensionality of the data, $\w \in \mathbb{R}^D$ is a weight vector, and $\epsilon$ is a noise term following the logistic distribution with zero mean and unit variance. 
Assuming $K$ categories, and a set of $K + 1$ thresholds $\theta_k \in \{\theta_0, \theta_1, \dots, \theta_{K}\}$ (ordered by $\theta_k < \theta_{k+1}$) one can assign a response $y$ according to the interval into which $y^*$ falls with the function $f_k: \mathbb{R} \xrightarrow{}  \{0,1\}$:

\begin{align}
    f_k(y^*) = {\begin{cases}
        1~~~~{\text{if}}~~\theta_ {k-1}<y^{*}\leq \theta _{k}\\
        0~~~~{\text{otherwise}}\end{cases}}
\end{align}

Three of the thresholds are fixed ($\theta_0=-\infty$, $\theta_1=0$ and $\theta_K=\infty$) to ensure that the process is identifiable \cite{kevin2012machine}. The probability over the categories is computed by integrating the probability mass that falls between the intervals

\begin{align}
    \label{eq:ordlogprob}
    P(y=k|\mathbf {x} ) 
        &=P(\theta _{k-1}<y^{*}\leq \theta _{k}|\mathbf {x} )\nonumber\\
        &=\sigma (\theta _{k}-\mathbf {w}^\top \mathbf {x} )-\sigma (\theta _{k-1}-\mathbf {w} ^\top \mathbf {x} )
\end{align}

\noindent where $\sigma(\cdot)$ is the logistic function (\ie cumulative distribution of the logistic distribution). 
The log-likelihood and its gradient with respect to the parameters ($\{\w, \theta_2, \theta_3, \dots, \theta_{K-1}\}$) are easily computed and can be optimised with standard optimisation techniques \cite{mccullagh1980regression}. Previous work presents an approach based on the Support Vector Machine and a dataset constructed by considering all the pairwise difference vectors \cite{herbrich1999support}. One of the main advantages of these models over simpler baselines (such as linear regression) is that the ordinal intervals are optimised during the learning routine and that the intervals can have arbitrary widths. 
It is important to understand that the primary assumption underpinning these models is that the data lies on a linear manifold, and in practice this is difficult to guarantee. Other approaches within the {threshold models} category include an adaptation of the online perceptron algorithm \cite{crammer2005online}, as well as an approach based on a generative model, which uses Gaussian Processes \cite{chu2005gaussian}.



\subsection{Decomposition Models}
\paragraph{Ordinary Binary Decompositions:} products of multiple binary models, or, single models capable of multiple-outputs. For example, in multi-class classification problems, one usually resorts to solving multiple smaller problems and then combining their predictions according to voting schemes such as \ac{OVO} or \ac{OVA}. Considering a problem with $K$ classes, in the former setting, one would need $K(K-1)/2$ {`small'} learners, while in the latter $K$ {`larger'} learners, where the distinction between small and large refers to the average size of the data they will be dealing with. OvA is also susceptible to the problem of class-imbalance. Based on the assumption of the ordering of the classes one could construct more developed voting schemes, that reflect his prior knowledge and reduce the computational complexity of the overall algorithm. Examples of such ordinal voting schemes include, \textit{one-vs-next}, \textit{one-vs-followers}, and decompositions based on \textit{Ordered Partitions} (see Section 3.2. in \cite{gutierrez2016ordinal}). 

These decompositions are closely related to the concept of \ac{ECOC}, which is used to reduce multi-class classification problems to combinations of binary tasks \cite{dietterich1994solving}. In this setting, every class is assigned to an `output code', which usually contains values in $\{-1, 0, +1\}^Q$. When considering multiple binary models, each of the $Q$ entries of this output code is generated by one of the models. The predicted class is the one whose output code is closer to the composition
of predictions. A similar line of work keeps the connection between classes and output codes, but instead of training one model per `bit', trains a model capable of multiple outputs on the whole code. In the simplest case this could amount to the output codes being of the form of the popular one-hot embedding, but \ac{ECOC} provides a framework for more delicate codes to be utilised, such as ones reflecting the prior knowledge of the classes being ordered. 

\paragraph{Nested Binary Classifiers:} 
A flexible ordinal model based on a decomposition of the label space can be produced with cascades of linear classifiers \cite{frank2001simple} by recasting the ordinal task into $K - 1$ independent binary classification problems. 
The $k$-th binary problem re-partitions the dataset into two groups; the first group consists of all instances whose label is less than or equal to the value $k$, and the second group consists of all instances with label greater than the value $k$. 

Using an equivalent rationale to that on Equation \eqref{eq:ordlogprob}, the probability distributions over each partition are unified into a probability distribution over the $K$ categories with the following equation

\begin{align}
    \label{eq:nbc}
    P(Y=k|\x) = P(Y > k-0.5|\x) - P(Y>k+0.5|\x)
\end{align}

\noindent with the base cases $P(y>0.5|\x)\triangleq1$ and $P(y>K|\x)\triangleq0$. 
The $K-1$ models are learnt independently, and only the two classifiers that `neighbour' the correct label are used in prediction. 

Although this model is simple and derived from an intuitive standpoint, it also carries several disadvantages. Firstly, the $K-1$ binary classifiers are learnt independently. While this brings gains in terms of concurrently learning each model it is unlikely that the final model will produce optimal decisions. 
Secondly, the mechanism for decision making shown in Equation \ref{eq:nbc} cannot guarantee consistency in classification and in general may require clipping and renormalisation for probabilistic predictions \cite{cardoso2007learning}, and this is particularly clear if one envisages an ordinal classification task when the data lies along complex or nonlinear data manifolds.  


In the taxonomy of algorithms presented in \cite{gutierrez2016ordinal}, the \textit{Ordinary Binary Decompositions} category has another sub-class of methods. Therein, a first group of methods takes advantage of the ordinal nature of the classes to devise clever decompositions, while the second group transforms the problem to a multi-target one, with ordinal encodings as targets. Models must be aware of the structured nature of the output space in order to take advantage of these encodings. 



\section{Methods}
\label{section:methods}


In this section we introduce our proposed technique for ordinal regression \acf{STORM}. We cast the ordinal regression task into a structured classification task. We use a simple encoding scheme for the labels which allows for a simple propagation of information through a CRF constructed from the label representation. Although classification methods in general are considered na\"{i}ve on the ordinal regression ontology (\cf \cite{gutierrez2016ordinal} and Section \ref{section:related_work:naive}) the proposed method is further developed (and hence not na\"{i}ve) since the label encoding deliberately captures several desirable properties of ordinal predictors. A key advantage of the application of \acp{CRF} to the encodings above is that one model is produced and optimised to produce outputs, in contrast to many approaches from the threshold and decomposition strand of the \ac{OR} ontology.

\begin{figure*}[t]
    \centering
    \includegraphics[width=\textwidth]{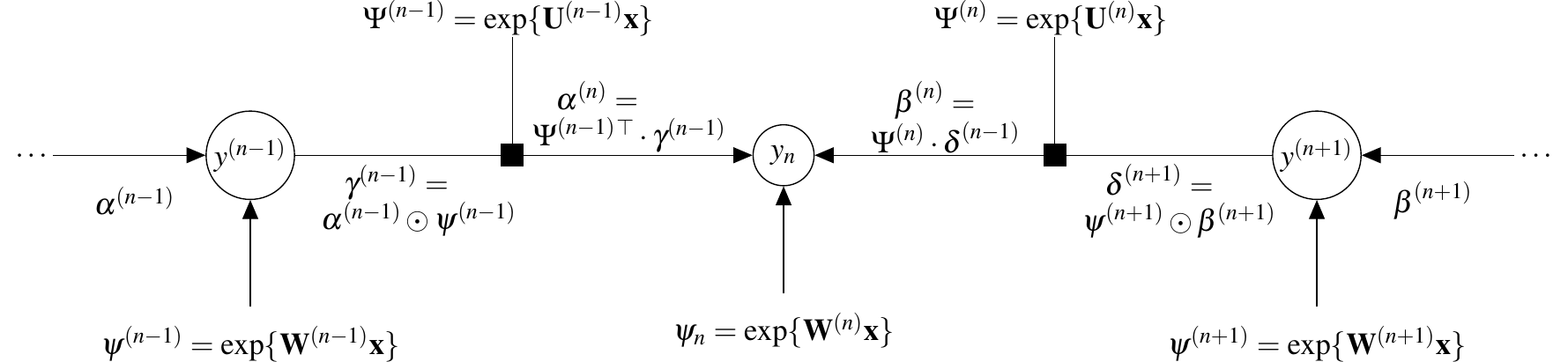}
    \caption{A graphical illustration of marginalisation process for CRFs. Notation is defined in Section \ref{sec:crf4or}.
    }
    \label{fig:crf}
\end{figure*}

%


\subsection{Label Encoding} 

A key enabler of the proposed approach is the symbiotic relationship between a bespoke encoding scheme for ordinal variables on one hand and the modelling framework that is used to infer and predict on the space of encoded labels on the other (next section). The encoding scheme that we use has previously been introduced for capturing resemblance measures for ordinal variables \cite[Ch. 8]{esposito2000similarity} but we believe we are the first that incorporate this representation directly into the modelling framework. 

We consider an ordinal problem as having $K$ categories, and our encoding scheme transforms these into a sequence of $K-1$ binary digits. The following function defines the value of the $k$-th bit of an encoded sequence:

\begin{align}
    \widehat{f}_K(\widehat{y}, k) = \begin{cases*}
      1 & if $k < \widehat{y}$\hspace{3em}$(1 \leq k \leq K - 1)$ \\
      0 & otherwise
    \end{cases*}
\end{align}

\noindent where the function subscript defines the support of the ordinal categories (\ie $K$) and $\widehat{y}$ ($1 \leq \widehat{y} \leq K$) is the `raw' (\ie un-encoded) label. As a concrete example, for $K=7$ and $\widehat{y}=4$, the encoded label $\y$ is given as: 

\begin{align}
    f_7(4) = \left( 1, 1, 1, 0, 0, 0\right)
\end{align}

\noindent where we have defined the new function

\begin{align}
    f_K(\widehat{y}) = \left(\widehat{f}_K(\widehat{y},1), \widehat{f}_K(\widehat{y},2), \dots, \widehat{f}_K(\widehat{y},K-1) \right)
\end{align}

To motivate this encoding scheme for \ac{OR}, consider two instances with $\widehat{y}^{(1)}=3$ and $\widehat{y}^{(2)}=5$ and their encoded values:

\begin{align}
    f_7(3) = \left( 1, 1, 0, 0, 0, 0\right) \\
    f_7(5) = \left( 1, 1, 1, 1, 0, 0\right) 
\end{align}

Recalling that these are the encoded representation of the labels of two instances, we can see that even though the raw labels are distinct that four bits of the encoded labels are of the same value. Thus, we can split the encoded labels into three virtual segments: 1) the first two bits which are positive and identical; 2) the final two bits which are negative and identical; and 3) the middle two bits which disagree and encode the intrinsic differences between the instances. In the next section we introduce a framework for modelling sequences of data that obey the constraints of the encoding and thus capture `shared' and `distinct' aspects of the encoded labels above. 

\subsection{Conditional Random Fields}
\label{sec:crf4or}


We utilise the language of probabilistic modelling and \acfp{CRF} in our setting. \acp{CRF} constitute a structured modelling framework \cite{lafferty2001conditional}, and in this section we motivate and introduce a generalisation of the traditional linear-chain \acp{CRF} for \ac{OR}. Linear-chain \acp{CRF} incorporate weight-sharing on all positions of a sequence since, for these models, the dynamics (\ie predictive response as a function of input) are stationary \cite{twomey2016need}. In other words the effect of one feature is equal at all positions of a sequence. This is a strong assumption, but in particular is inappropriate with our encoded labels since a feature may need to have diminishing (or increasing) responses depending on the position of the sequence. 
For the remainder of this section we assume the reader has familiarity with \acp{CRF} and recommend the following as an introducton: \cite{sutton2012introduction}. 


To overcome this incompatibility, we use the \ac{CRF} framework with but importantly without weight sharing. We have a dataset that consists of $N$ observations of dimensionality $D$, \ie $\X \in \mathbb{R}^{N \times D}$. With $K$ ordinal quantities the encoded labels are $\Y \in \{0,1\}^{N\times(K-1)}$. In order to simplify mathematical notation for the remainder of this section we focus on one particular example/label pair ($\x$, $\y$) which can be considered as the $i$-th row of $\X$ and $\Y$ respectively. Of critical importance for this method is the fact that the label has been mapped from the `one-of-K' encoding to the `up-to-k' encoding, and hence the space of labels (and predictions) have become a sequence of binary variables for every instance. Although this might be viewed as an unnecessary complication (since no new information is introduced) we will later see the value that is introduced by this encoding. 


\subsubsection{CRFs for OR}

\acp{CRF} yield structured predictions over graphs. In our setting, the graph consists of $K-1$ nodes with $K-2$ edges linking the nodes together in a chain. Each node (indexed by $n$) contains its own set of weights as does each edge (indexed by $e$). We follow standard potential and marginalisation methods from the \ac{CRF} literature. First, node and edge potentials are computed. The $n$-th node potential is given by

\begin{align}
\label{eq:node_potentials}
    \psib^{(n)} = \exp\{\W^{(n)} \x\}
\end{align}

\noindent where $\x \in \mathbb{R}^D$ is the feature vector and $\W^{(n)}\in\mathbb{R}^{2 \times D}$ is the weight vector associated with the $n$-th node, and $\psib^{(n)} \in \mathbb{R}^{2} ~ \forall~n$. To simplify notation we assume that a `bias feature' with constant value of 1 is contained in the feature vector $\x$. Similarly the potential of the $e$-th edge is given

\begin{align}
\label{eq:edge_potentials}
    \Psib^{(e)} = \exp\{\mathbf{U}^{(e)} \x\}
\end{align}

\noindent where $\mathbf{U}^{(e)} \in \mathbb{R}^{2 \times 2 \times D}$ is the weight tensor associated with the $e$-th edge of the model and multiplication takes place on the outermost dimension, and $\Psib^{(e)} \in \mathbb{R}^{2\times2} ~\forall~ e$.



Inference can be performed with standard message passing which can efficiently be computed with the forward-backward dynamic program.  The $n+1$ forward vector is given by

\begin{align}
\label{eq:fward}
    \alphab^{(n+1)} = \displaystyle{\Psib^{(n)\top}} \gammab^{(n)}
\end{align}

\noindent where $\top$ represents the matrix transpose,  $\gammab^{(n)} \triangleq \alphab^{(n)} \odot \psib^{(n)}$ and $\odot$ represents the Hadamard product. The $n-1$ backward vector is calculated similarly with

\begin{align}
\label{eq:bward}
    \betab^{(n-1)} = \Psib^{(n)} \deltab^{(n)}
\end{align}

\noindent where $\deltab^{(n)} \triangleq \psib^{(n)} \odot \betab^{(n)}$, and the base cases for the forward and backward vectors are $\alphab^{(0)} \triangleq \boldsymbol{1}$ and $\betab^{(K)} \triangleq \boldsymbol{1}$. Note, marginalisation is often performed in the log domain with the log-sum-exp function for numerical stability but identical marginal distributions are achieved to those above.

It can be shown that the forward and backward vectors yield sufficient information for exact marginal probability estimation \cite{sutton2012introduction} and the probability of the $n$-th position of the label is given by 

\begin{align}
    P(\y_n) = \alphab^{(n)} \odot \psib^{(n)} \odot \betab^{(n)} / Z
\end{align}

\noindent where $Z$ is the global normaliser of the sequence that can be calculated at any position, $Z = \mathbf{1}^\top (\alphab^{(n)} \odot \psib^{(n)} \odot \betab^{(n)}) $, and the probability across the $n$-th edge is

\begin{align}
\label{eq:transprob}
    P(\y_n, \y_{n+1}) = \gamma^{(n)} \odot \Psib^{(n)} \odot \delta^{(n+1)\top} / Z
\end{align}

Figure \ref{fig:crf} illustrates the inference procedure along a graph. Three nodes are shown here, and each of the intermediate quantities introduced earlier are shown. 

\subsubsection{Learning}
Optimisation is performed by minimising the negative logarithm of the likelihood, \ie 

\begin{align}
    \mathcal{NLL} = -\sum_{n=1}^N \log P(\Y_n | \X_n, \Theta)
\end{align}

\noindent where $\Theta = \{ \W^{(1)}, \W^{(2)}, \dots, \W^{(K-1)}, \mathbf{U}^{(1)}, \mathbf{U}^{(2)}, \dots, \mathbf{U}^{(K-2)} \}$ is the set of model parameters. It is easy to show that the gradient of the $i$-th element of the $n$-th weight vector is: 

\begin{align}
    \frac{\delta \mathcal{NLL}}{\delta \W^{(n)}_{i}} =\frac{1}{N} \sum_{j=1}^N \left( P\left(\Y_{j,n} = i\right|\X_j, \Theta) - \mathbb{I}\{\Y_{j,n} = i\}\right) \X_j
\end{align}

\noindent where $\mathbb{I}\{\cdot\}$ is the identity function, and derivation of the above follows similar methodology for other log-linear models \cite{lafferty2001conditional,sutton2012introduction} and very similar expressions can be produced to produce gradients with respect to the edge weights $\mathbf{U}^{(e)}_{i,j}$. Standard gradient-based optimisation techniques can be used to minimise the negative log likelihood, \eg L-BFGS. 


It is interesting to note that even though log loss is optimised here that the structure of the labels can be seen to be functionally related the absolute error between labels and predictions. One can view this either as a hybrid loss function or that the proposed methodology implicitly applies misclassification costs owing to the structure of the encoded label space. 


\subsubsection{Comments on the Model}
%
Since many aspects of this model are unexplored in the field of \ac{OR} we take a moment to comment on some aspects of this model in this setting

\textit{Edges:} We interpret the edges of the model as driving the `transitions' between two adjacent encoded bits. In more traditional sequence learning settings, including natural language processing, is it very typical to direct bespoke features for the edges only. We ascribe a similar interpretation of the edges in our setting, \ie the $n$-th edge primarily drives whether the $n$-th bit of the encoded label is sustained or transitions whereas the node weights drive the basic identification of categories. 

\textit{Predictive Distribution:} \acp{CRF} facilitate several methods for producing predictions: forward filtering, Viterbi path, marginal probability distribution of the sequence \cite{kevin2012machine}. Although in this work we consider the Viterbi path, we acknowledge that existing literature exists that suggests other predictive functions to be used when optimising for different performance metrics. 

\textit{Errors:} Not all paths are permissible with our encoding scheme, with $0\rightarrow1$ transitions in particular being forbidden. Several approaches can be incorporated to produce predictions consistent with the encoding. Firstly, one can explicitly forbid illegal transitions by setting $\Psib^{(e)}_{0,1}=0$. However, in our experimentation we recognised some evidence that there is a correspondence between invalid predictions and outlying data. This work is ongoing.

\subsubsection{Implementation Notes}
Due to the specific nature of our problem, some operations can be vectorised to increase learning and inference time. Since all sequences are of the same length ($K-1$) the message passing procedure can be vectorised across all instances. In so doing forward messages will be passed from position $n$ to $n+1$ across all instances. Similarly, backward messages can be passed in a similar vectorised manner. This is often not possible due to the fact that most sequence learning problems have instances of different length. 

Furthermore, if the carnality of the ordinal problem is small (in our experiments less than $K < 30$) inference can be performed the exponential domain without re-normalisation without noticeable loss of fidelity in probability estimates. This yields significant gains in terms of the computational time since neither the logarithm or exponential functions are used for marginalisation. 

\section{Experiments}

\label{section:experiments}

\subsection{Models}

We compare four different models that are linear in their parameters. We only consider linear models so that we can compare the proposed method with baselines in the `natural' data representations. Practitioners that wish generalise this work and consider nonlinear predictors may incorporate kernel functions (polynomial, for example) or explicitly parameterise nonlinear representations with deep network architectures. 
Hence, we consider the following linear models only: 

\begin{enumerate}
    \item \acf{OL}; 
    \item \acf{NEST}; 
    \item \acf{LR}; and 
    \item \acf{STORM}. 
\end{enumerate}


These models are all log-linear and regularisation was performed on the weight parameters and we perform crossvalidation over the $\ell_2$ norm of the parameters. We select the regularisation parameter on the training set using 5-fold cross validation. 


\subsection{Datasets}

Table \ref{tab:datasets} shows the characteristics of the datasets considered in the empirical evaluation in this paper. The table presents four categories of datasets (synthetic, UCI, large and health) and these are explained in the subsequent subsections. 


\begin{table}[h]
    \centering
    \caption{The datasets that are considered in this work.}
    \begin{tabular}{cllllc}
         \toprule
         {} & {Dataset} & {Features} & {Train} & {Test} & {K} \\
         \midrule
         \multirow{4}{*}{\rotatebox[origin=c]{90}{{\sc Synthetic}}}
            & \linear{}                 & 2 & 100 & 1000 & 5 \& 10 \\
            & \jagged{}                 & 2 & 100 & 1000 & 5 \& 10 \\
            & \circle{}                 & 2 & 100 & 1000 & 5 \& 10 \\
            & \spiral{}                 & 2 & 100 & 1000 & 5 \& 10 \\
         \midrule
         \multirow{9}{*}{\rotatebox[origin=c]{90}{{\sc UCI}}}
            & Diabetes                   & 2  & 30   & 13   & 5 \& 10 \\
            & Pyrimidines               & 27 & 50   & 24   & 5 \& 10 \\
            & Triazines                 & 60 & 100  & 86   & 5 \& 10 \\
            & Wisconsin                 & 32 & 130  & 64   & 5 \& 10 \\
            & Machine CPU               & 6  & 150  & 59   & 5 \& 10 \\
            & AutoMPG                   & 7  & 200  & 192  & 5 \& 10 \\
            & Boston Hous            & 13 & 300  & 206  & 5 \& 10 \\
            & Stocks              & 9  & 600  & 350  & 5 \& 10 \\
            & Abalone                   & 8  & 1000 & 3177 & 5 \& 10 \\
         \midrule
         \multirow{7}{*}{\rotatebox[origin=c]{90}{{\sc Large}}}
            & Bank 1             & 8  & 50   & 8142   & 5 \& 10 \\
            & Bank 2             & 32 & 75   & 8117   & 5 \& 10 \\
            & CompAct1        & 12 & 100  & 8092   & 5 \& 10 \\
            & CompAct2        & 21 & 125  & 8067   & 5 \& 10 \\
            & Cali Hous         & 8  & 150  & 15490  & 5 \& 10 \\
            & Census1            & 8  & 175  & 16609  & 5 \& 10 \\
            & Census2            & 16 & 200  & 16584  & 5 \& 10 \\
         \midrule
         \multirow{2}{*}{\rotatebox[origin=c]{90}{{\sc AD}}}
            & \dementiabank{}           & 1605 & 200 & 169 & 4 \\
            & \casas{}                  & 278  & 200 & 118 & 5 \\
         \bottomrule
    \end{tabular} 
    \label{tab:datasets}
\end{table}

\subsubsection{Synthetic}
For our synthetic experiments, we project data onto the four following data manifolds: \begin{inparaenum}
    \item \linear{}; 
    \item \jagged{}; 
    \item \circle{}; and 
    \item \spiral{}.
\end{inparaenum} 
These data manifolds lie in 2D spaces, and we illustrate the predictive distributions of all models visually in order to understand the strengths and limitations of each model. 
Empirical validation is performed with $K=5$ and $K=10$. These datasets are shown visually in our results and discussion. 

\subsubsection{UCI \& Large}
We follow \cite{chu2005gaussian} with two categories of datasets. The the following datasets from the UCI machine learning repository: \autompg, \diabetes, \abalone, \bostonhousing, \machinecup, \pyrimidines, \stocksdomain, \triazines, and \wisconsin. 
Although many of these datasets are used to understand regression models, we incorporated equal-frequency binning on these datasets so that they can be used in ordinal tasks. 
Empirical validation is performed with $K=5$ and $K=10$. We also consider a second (larger) set of data that was also introduced in \cite{chu2005gaussian} as the `large' dataset.

\subsubsection{Healthcare}
Finally, we also evaluate our model on two \ac{AD} datasets.
\dementiabank{} \cite{dementiabank} is a longitudinal dataset of multimedia interactions for the study of communication in dementia. The dataset contains transcript and audio files from interviews between patients and clinicians, and covers a range of diagnostic tests in mental health, such as Alzheimers Dementia, Parkinsons, and mild cognitive impairment. The transcripts and audio files were gathered as part of a larger protocol administered by the Alzheimer and Related Dementias Study at the University of Pittsburgh School of Medicine. We use the \dementiabank{} dataset in an ordinal regression setting to model the various stages of progression of \ac{AD}: cognitively healthy, possible dementia, probable and dementia.

The \ac{CASAS} research group produce models and datasets for smart-home behaviour modelling. Their datasets consist of sensor data (including \ac{PIR}, temperature, door and object sensors) derived from naturalistic living in a \ac{SH} environment. The `cognitive assessment activity dataset' \cite{dawadi2013automated,crandall2012smart,dawadi2011approach} consists of approximately 400 participants performing several \acp{ADL} and \ac{IADL} in the \ac{SH}. Cognitive clinicians graded the activities were graded by domain experts on a range of 1-5, and predicting the assigned grade from sensor data is the task that we investigate here.

\subsection{Performance Evaluation}
\label{sec:perfeval}
All datasets are partitioned randomly into 20 folds on the `synthetic', `UCI' and `healthcare' datasets (\cf Table \ref{tab:datasets}). Following the protocol of \cite{chu2005gaussian} we also performed 100 randomised splits for the `large' datasets.  Model hyperparameters are selected with 5-fold cross validation on the training set, and the selected parameters are used for performance evaluation on the test set. We follow \cite{baccianella2009evaluation,gutierrez2016ordinal} in our evaluation metrics and use macro-averaged 0/1 loss, \ac{MAE}. 

\begin{figure}
    \centering
    \includegraphics[width=0.6\linewidth]{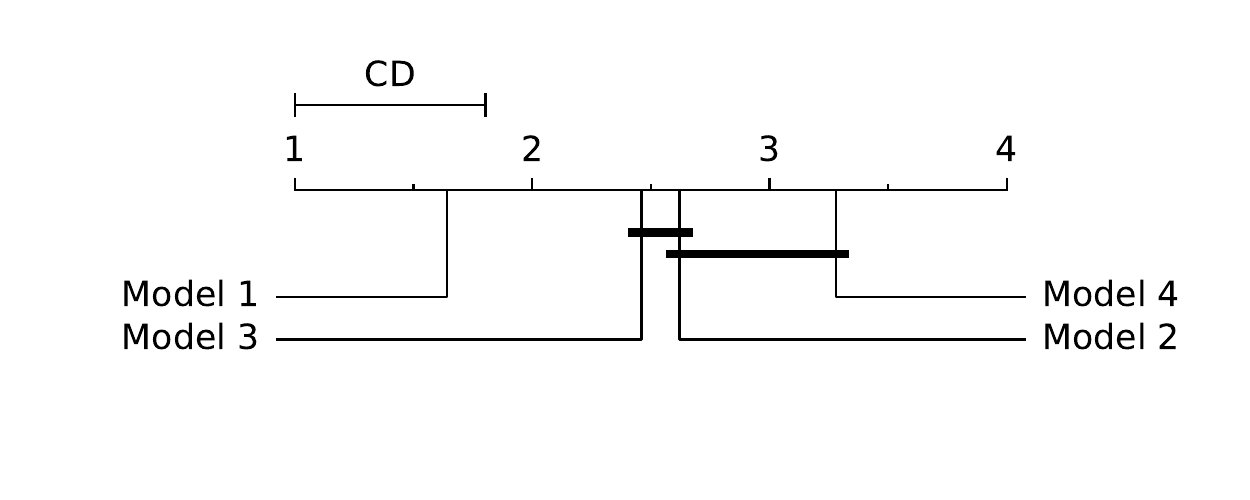}
    \caption{Example critical difference diagram.}
    \label{fig:excd}
\end{figure}

Additionally, significance of results is reported with the Wilcoxon's signed rank test \cite{benavoli2016should} at a (fairly stringent) significance level of $\alpha=0.01$. We illustrate the statistical significance with critical difference diagrams \cite{demvsar2006statistical} that are for the understanding of statistical significance when multiple classifiers are compared over multiple datasets. An example is shown in Figure \ref{fig:excd}. Four classifiers are shown here (Model 1, Model 2, Model 3, and Model 4) and the average rank of each is marked on the number line. The groups of algorithms whose results are not statistically different are connected together with a heavy horizontal line, \ie the difference between Models 2 and 3 is not statistically significant, whereas the difference between Models 1 and 2 is.

\section{Results}
\label{section:results}

In this section we present and discuss the results from the synthetic, UCI, large and healthcare datasets and conclude by discussing the complete results together.


\subsection{ Synthetic Datasets}

\subsubsection{Results}
\begin{figure*}
    \centering
    \subfigure[\circle{} dataset with 5 categories]{\label{fig:nonlin1}\includegraphics[width=\linewidth]{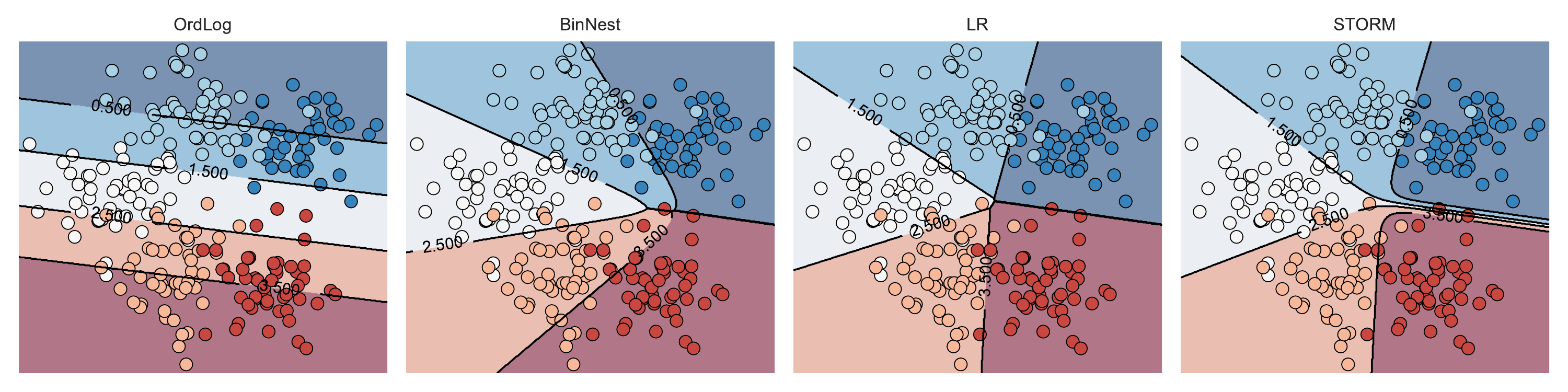}}\\
    \subfigure[\spiral{} dataset with 5 categories]{\label{fig:nonlin2}\includegraphics[width=\linewidth]{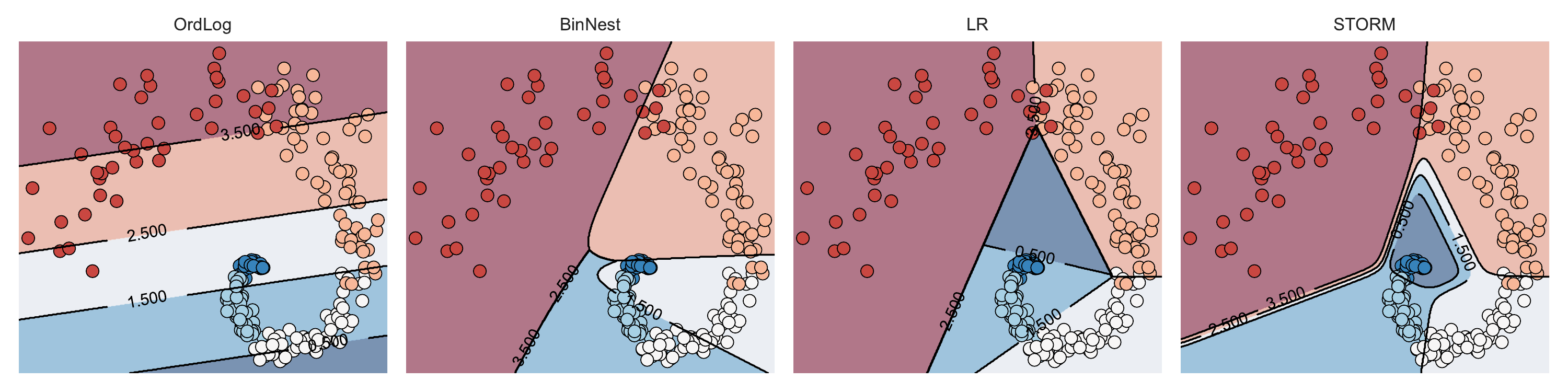}}
    \caption{
    Predictions from baseline and proposed ordinal 
    \circle{} and \spiral{} datasets (Figures \ref{fig:nonlin1} and \ref{fig:nonlin2} datasets).
    }
    \label{fig:synthetic}
\end{figure*}

We first present our results on synthetic datasets visually since these datasets are in two dimensions. 
The upper two subfigures of Figure \ref{fig:synthetic} present the results from the four classifiers considered (\ac{OL}, \ac{NEST}, \ac{LR}, and \ac{STORM}) on the \circle{} and \spiral{} (we show the \linear{} and \jagged{} predictions in the supplementary material). The dots represent instances in a two dimensional space, and the fill colour of each depicts the ground-truth label; darkest blue representing class 1 and darkest red representing class K. Additionally, the background colour in these figures represents the predicted ordinal quantities obtained from each model. The colour scheme is shared between the background and fill colours. 


Figure \ref{fig:nonlin1} and \ref{fig:nonlin2} show the predictions obtained when the ordinal data lies on a \circle{} and \spiral{} manifolds respectively on the 5-category dataset. Clearly, due to the limitations of the \ac{OL} model it cannot perform optimally here. Additionally the \ac{NEST} model does not adapt to the dynamics of the data manifold in this setting either since the greedy learning routine cannot resolve these manifolds (particularly with the \spiral{} dataset). The non-ordinal \ac{LR} model and the proposed \ac{STORM} are better able to adapt to the challenges with these data manifolds, with \ac{STORM} adapting most efficiently. 


We have observed noteworthy behaviour with the \ac{STORM} on all synthetic experiments, namely that the space of low-valued predictions tend to be `consumed' the domain of higher-valued predictions. 
This phenomenon is illustrated clearly in Figure \ref{fig:nonlin2} (right) with the spiral dataset and the \ac{STORM} model, but can also be observed in Figures \ref{fig:nonlin1}. 
This is achieved due to the encoding of the labels and is a fundamental property of \ac{STORM}. However, this is also a feature of many \ac{OL} models, but cannot be guaranteed by the other baselines we consider, \eg \ac{NEST} or \ac{LR}.

\begin{figure}
    \centering
    \subfigure[Macro-averaged 0/1 loss over synthetic datasets]{\label{fig:synthetic_01}
        \includegraphics[width=\linewidth]{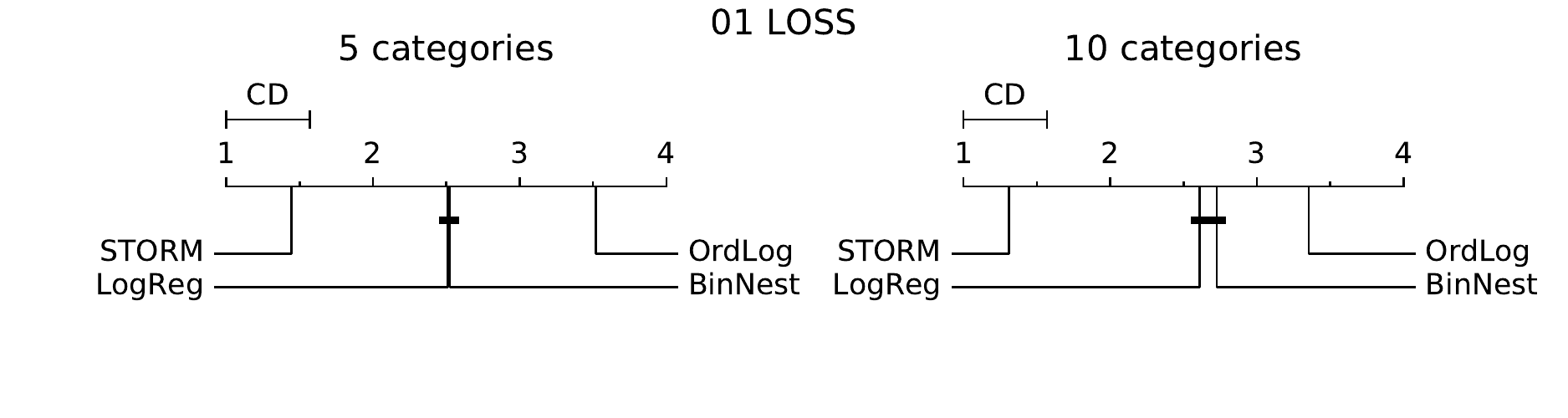}} 
    \\
    \subfigure[Macro-averaged mean absolute loss over synthetic datasets]{\label{fig:synthetic_mae}
    \includegraphics[width=\linewidth]{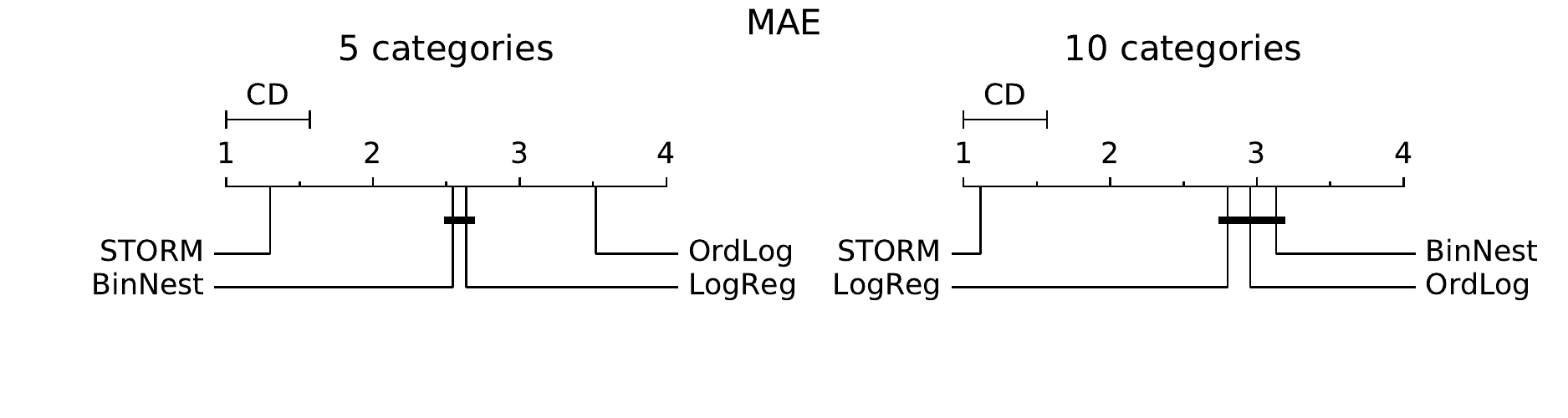}} \\
    \caption{
    Critical difference diagrams for synthetic datasets over the 20 train/test permutations.
    }
    \label{fig:cd_synthetic}
\end{figure}

Following \cite{baccianella2009evaluation,gutierrez2016ordinal}, we quantified performance using two metrics: macro-averaged mean absolute error and macro-averaged mean 0/1 loss. Figure \ref{fig:cd_synthetic} shows the critical difference diagram \cite{demvsar2006statistical} for the mean zero-one loss (Figure \ref{fig:synthetic_01}) and mean absolute error (Figure \ref{fig:synthetic_mae}). (For a description on how to read and interpret critical difference diagrams we refer the reader to Section \ref{sec:perfeval} and more generally to \cite{demvsar2006statistical}.) We can see from this figure that the proposed approach is ranked best and that its performance is significantly better than those of the baselines on all performance metrics considered. 


\subsubsection{Versatile Queries}

Since the language of probabilistic graphical models underpin the proposed method, \ac{STORM} may be queried in a variety of ways. In particular, here we will demonstrate how non-standard queries can be made by visualising the predictive distribution on an edge transition. 

\begin{figure}
    \centering
    \includegraphics[width=0.9\linewidth]{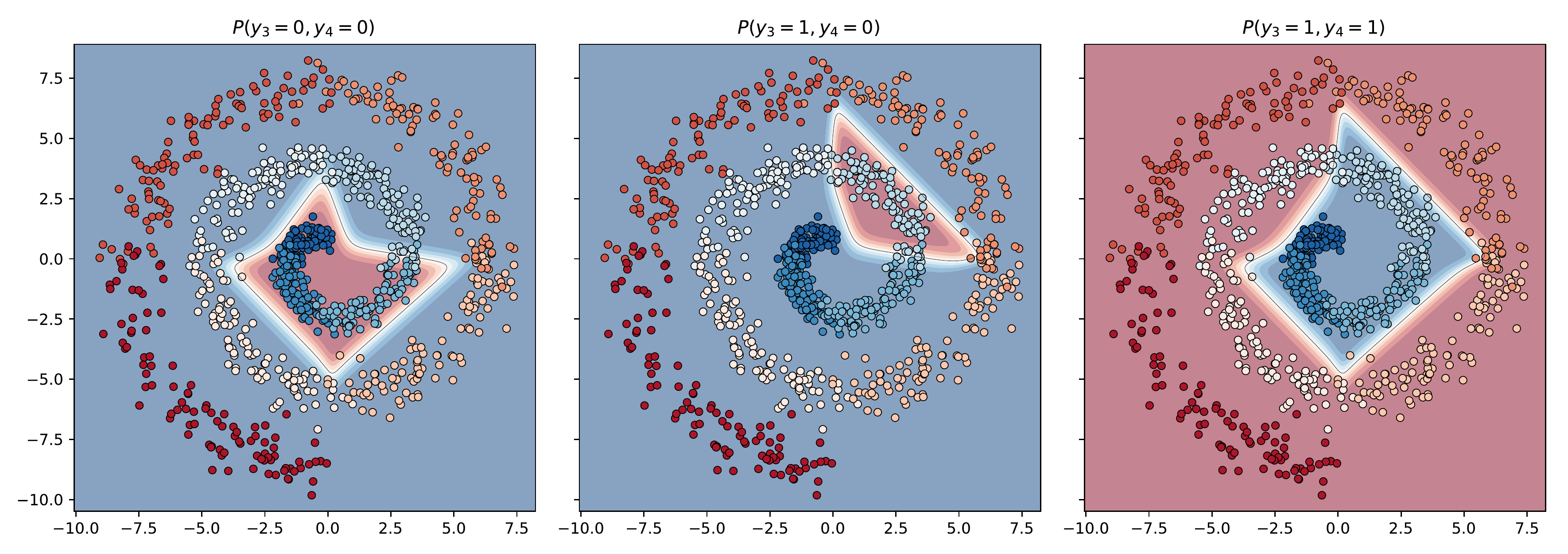}
    \caption{$P(y<4)$ (left), $P(y=4)$ (middle), and $P(y>4)$ (right).}
    \label{fig:queries}
\end{figure}

The probability distribution over the transition between the $i$-th and $(i+1)$-th positions is given by marginalising over all other positions, \ie

\begin{align}
    P(Y_i, Y_{i+1}) &= \sum_{Y_1}\sum_{Y_2} \dots \sum_{Y_{i-1}} \sum_{Y_{i+2}} \dots \sum_{Y_{K-1}}\sum_{Y_K} P(Y_1, Y_2 \dots Y_K)
\end{align}

\noindent and this can efficiently computed with forward and backward vectors (see Equation \ref{eq:transprob}).
%
%
Figure \ref{fig:queries} depicts the probability distribution over the transition between positions 3 and 4 on a variation of the 10-category \spiral{} dataset. Regions shaded in blue and red represent regions of low and high predicted probability respectively. The left figure shows $P(Y_3=0, Y_4=0)$, the middle figure shows $P(Y_3=1, Y_4=0)$, and the right figure shows $P(Y_3=1, Y_4=1)$. These probability distributions can be interpreted as $P(Y<4)$ in the left figure captures, $P(Y=4)$ in the middle captures and the right figure shows $P(Y>4)$. To understand why, we can consider at the third and fourth tags of encoded labels for several labels, and observe that the third and fourth tags for $\widehat{y}<4$ are both 0, for $\widehat{y}>4$ are both 1, and for $\widehat{y}=4$ we observe the pair $(1,0)$ corresponding to Figure \ref{fig:queries}.  

\begin{figure}
    \centering
    \subfigure[Linear features.]{\label{fig:vq1}\includegraphics[width=0.45\linewidth]{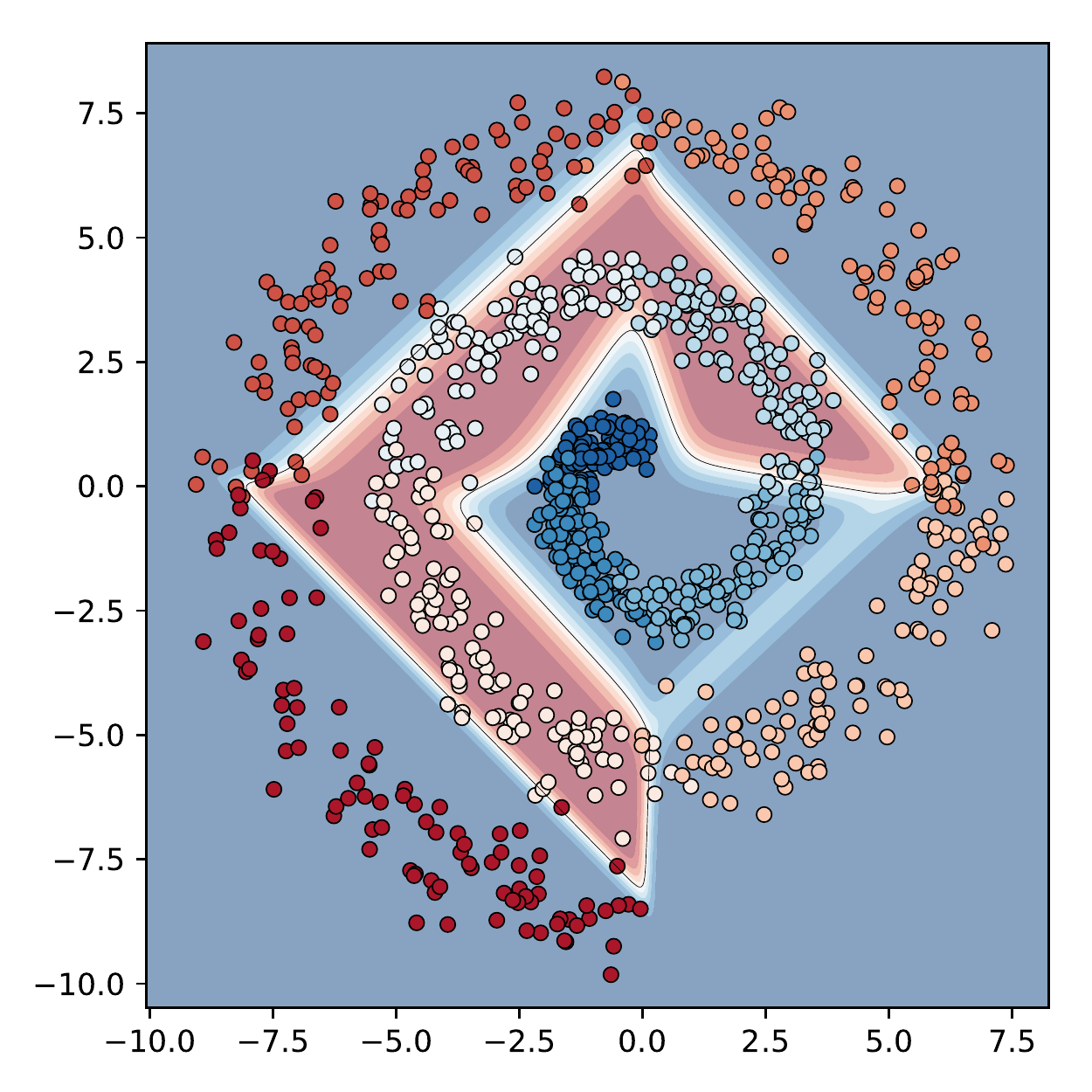}} 
    \subfigure[Nystr\"{o}em features.]{\label{fig:vq2}\includegraphics[width=0.45\linewidth]{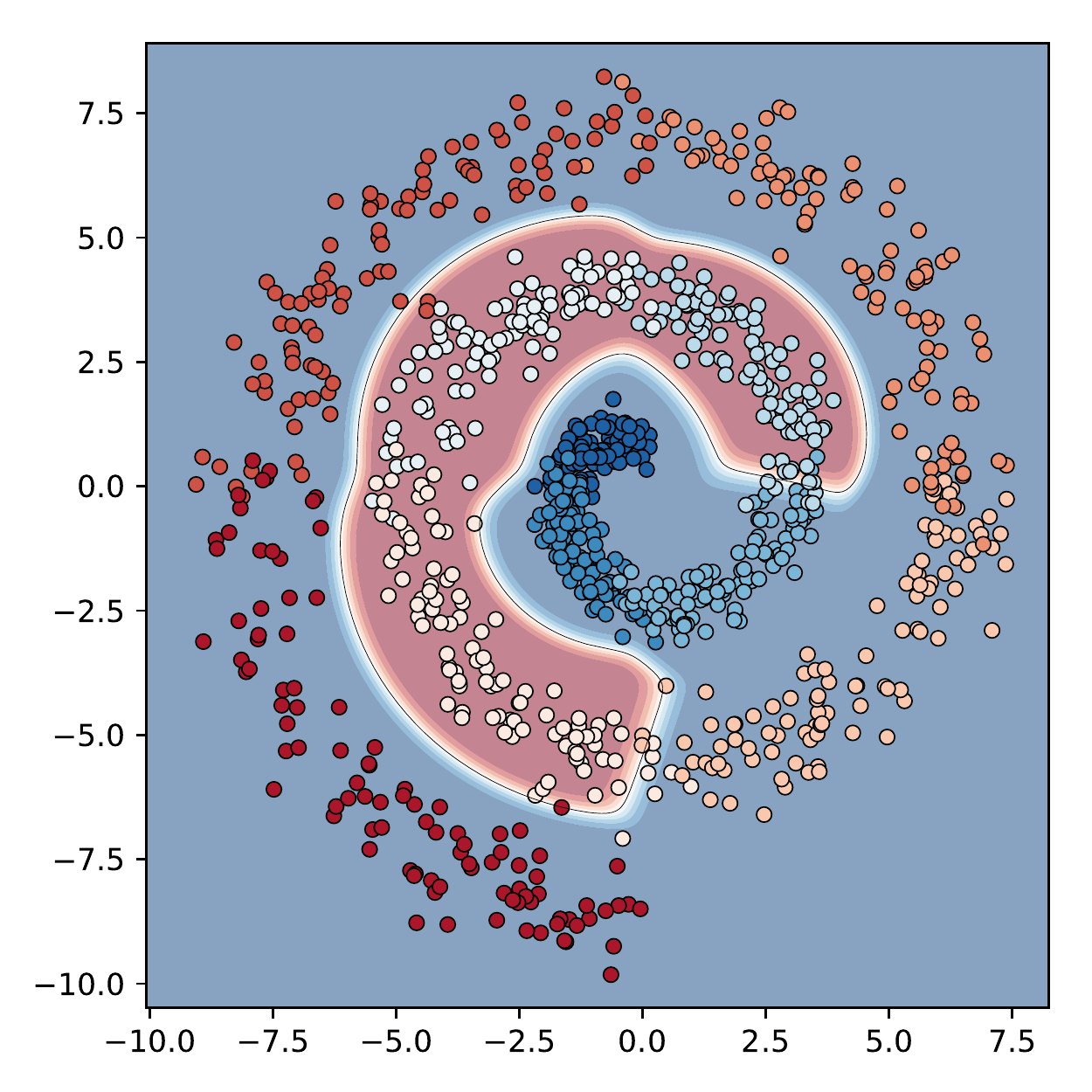}} 
    \caption{Versatile querying of the \storm{} on \spiral{}. 
    }
\end{figure}

Although the model itself is linear in its parameters the predictive distribution has adapted to the nonlinear data manifold. In settings with large $K$ (\ie many ordinal categories) one can easily execute more general queries (\eg $P(4 \leq Y < 7)$). As discussed in Section \ref{section:introduction}, this is a common task in clinical settings, \eg \ac{AD} patients will first be graded on a large scale before these are reduced into important intervals. The predictive distribution of $P(4 \leq Y \leq 7)$ may be indicative of a particular grade (\eg `moderate' \ac{AD}) and can be computed in our model. We demonstrate this visually for the a variant of the spiral dataset in Figure \ref{fig:vq1}. 
Although the focus of this work is on linear settings, we demonstrate the effect of Nystr\"{o}em kernel approximation \cite{williams2001using} of the \ac{RBF} kernel in Figure \ref{fig:vq2}. We notice that the nature of the data manifold is better captured with this representation and the predictive distribution curves alongside the manifold.

\subsection{Predictive Performance on UCI Datasets}

\begin{figure}
    \centering
    \subfigure[Macro-averaged 0/1 loss over UCI datasets]{\label{fig:uci_01}
    \includegraphics[width=\linewidth]{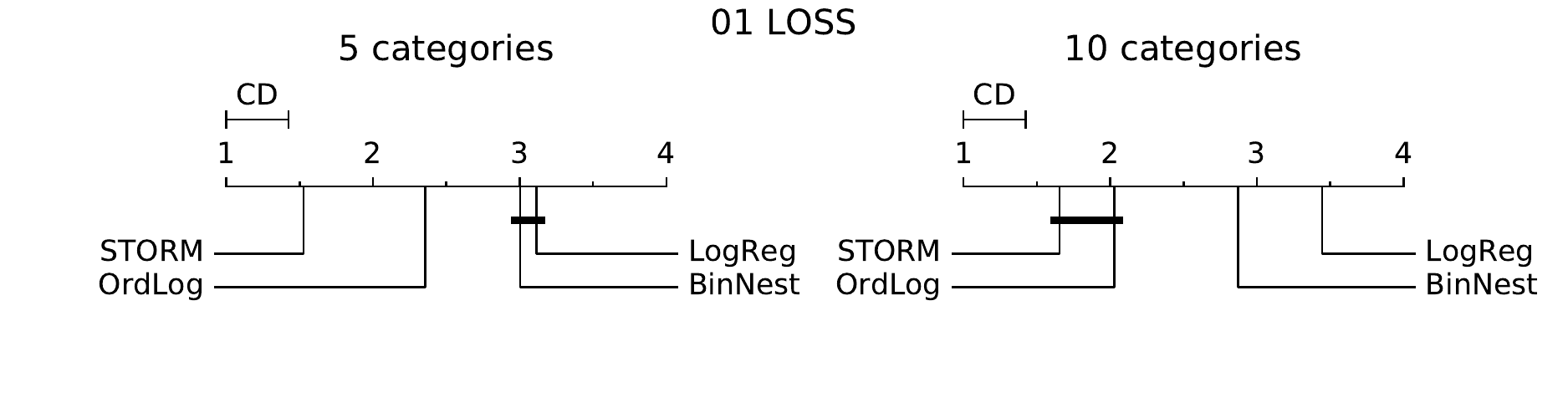}} 
    \\
    \subfigure[Macro-averaged mean absolute loss over UCI datasets]{\label{fig:uci_mae}
    \includegraphics[width=\linewidth]{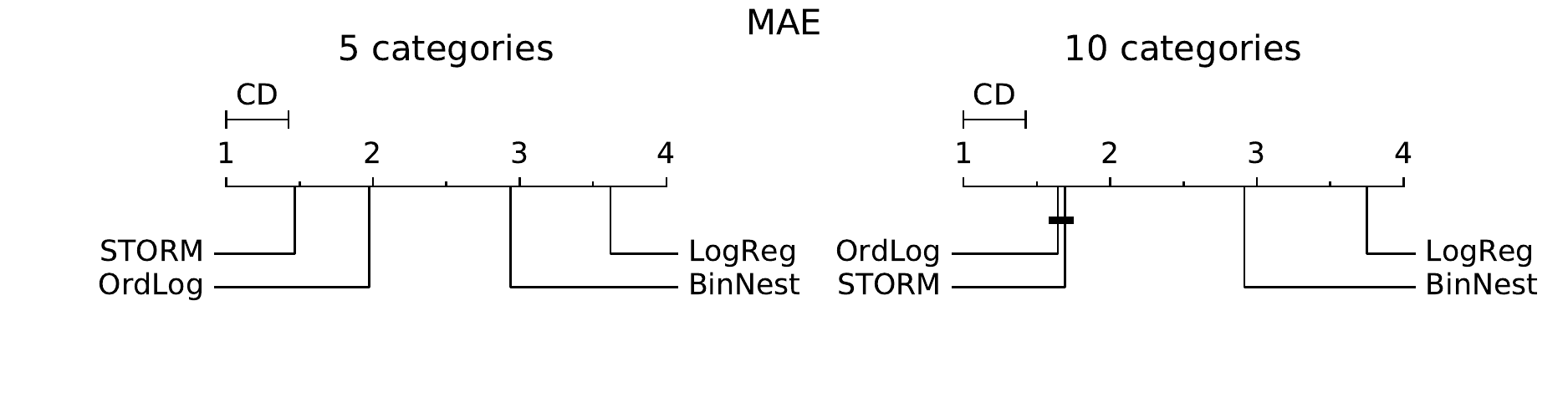}} \\
    \caption{
    Critical difference diagrams for UCI datasets.
    }
    \label{fig:real}
\end{figure}

Predictive performance was also evaluated on several datasets from the UCI machine learning datasets repository \cite{asuncion2007uci}. Figure \ref{fig:real} presents the critical difference diagrams for the 0/1 loss (Figure \ref{fig:uci_01}) and mean absolute error (Figure \ref{fig:uci_mae}) over 5 (left) and 10 (right) categories. Figure \ref{fig:real} illustrates that the \ac{STORM} model is the best performing model over all metrics with 5 categories, and its performance is significantly better on all metrics with the sole exception of mean squared error. 
Figure \ref{fig:real} also shows that other models are competitive with \ac{STORM} on the 10-category datasets. \ac{STORM} is never significantly less performant than the winning model, but is significantly better than \ac{LR} and \ac{NEST} baselines. 

\begin{figure}
    \centering
    \subfigure[Macro-averaged 0/1 loss over large datasets]{\label{fig:large_01}
    \includegraphics[width=\linewidth]{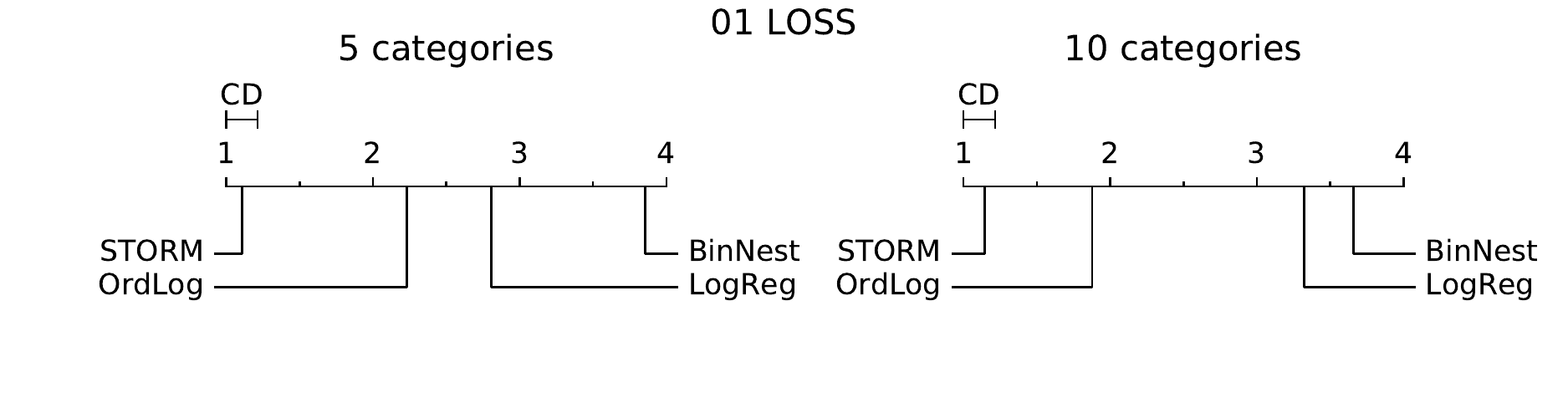}} 
    \\
    \subfigure[Macro-averaged mean absolute loss over large datasets]{\label{fig:large_mae}
    \includegraphics[width=\linewidth]{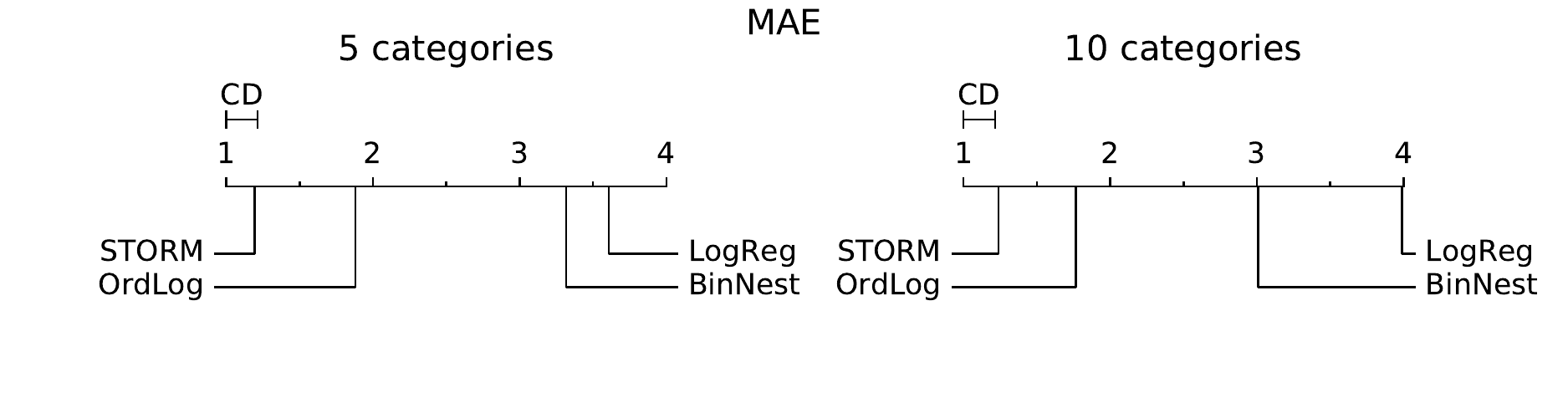}} \\
    \caption{
    Critical difference diagrams for large datasets.
    }
    \label{fig:large}
\end{figure}

We test the performance of \storm{} with larger datasets (in terms of number of instances and features) with the `large' dataset from \cite{chu2005gaussian}, and the results of these are shown in Figure \ref{fig:large}. These experiments were repeated over 100 randomised folds with 5 and 10 categories. \ac{STORM} significantly outperforms baseline approaches. 

\subsection{Healthcare Datasets}

Finally, we present results on the healthcare datasets. For the \ac{CASAS} dataset we were unable to produce the same feature representations that were used in the original paper since some of the data is withheld to preserve anonymity. We extracted the duration of the activity, the number of unique sensors, the most commonly triggered sensor, and the number of sensors from each category (presence, door, object \etc) that were triggered. The task of this dataset is to estimate the `incompleteness' of an \ac{AD} with 5 meaning the task was not completed and 1 good completion. 

With \dementiabank{}, we analysed the transcripts of the interviews conducted with the participants and defined an ordinal task on the following order: cognitively healthy, possible dementia, probable dementia, dementia. The transcripts also included annotations of pausing and verbal disfluency. Data representation consisted of counting occurrences and normalising features. 

\begin{figure}
    \centering
    \subfigure[0-1 loss on healthcare datasets.]{\label{fig:health:01}
        \includegraphics[width=0.45\linewidth]{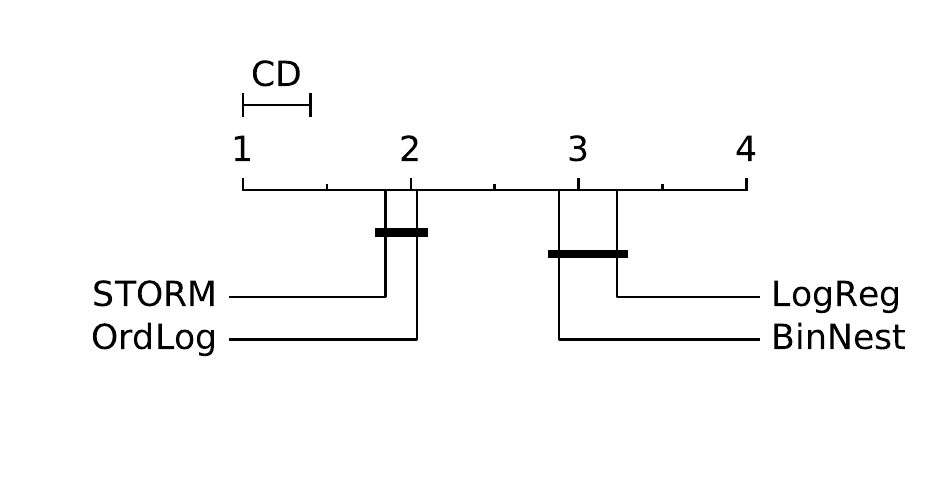}
    } 
    ~
    \subfigure[MAE on healthcare datasets.]{\label{fig:health:mae}
        \includegraphics[width=0.45\linewidth]{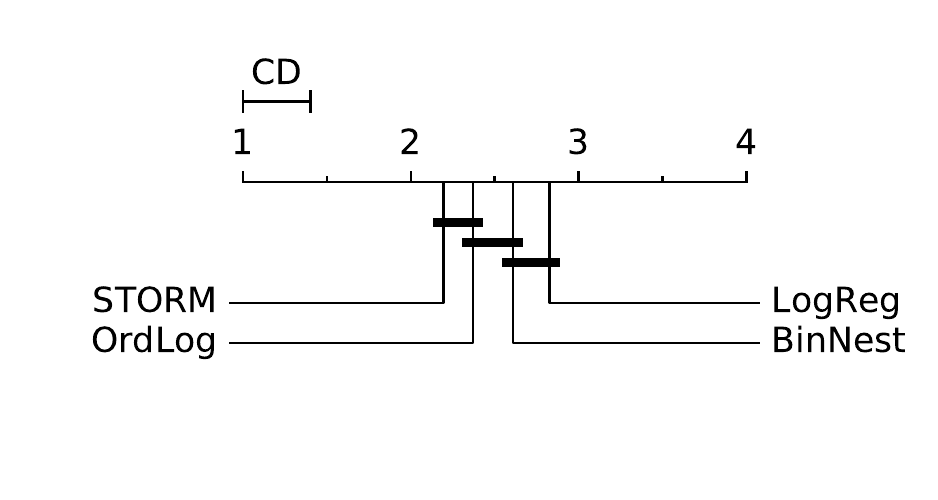}
    } 
    \caption{
    Critical difference diagrams on the healthcare.
    }
    \label{fig:healthcare}
\end{figure}

Figure \ref{fig:healthcare} presents the critical difference diagrams for the healthcare datasets. Note, that in all cases the critical difference in these figures is larger than in the synthetic and UCI datasets due to the smaller number of datasets here. These experiments have produced a much more competitive set of results with no one model consistently out-performing the others in a statistically meaningful manner. The \storm{} model is the best performing model over all tests conducted, even though its performance is not significantly better than ordinal regression. 

\begin{figure}
    \centering
    \includegraphics[width=0.5\linewidth]{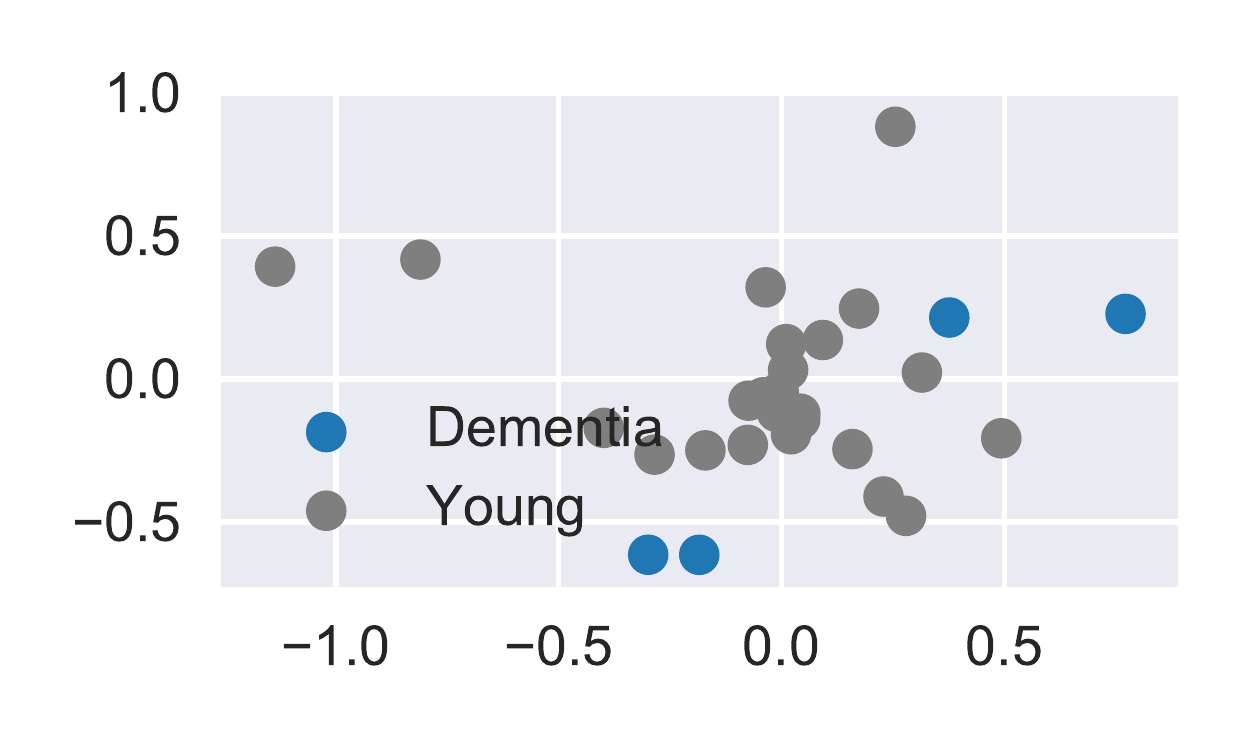}
    \caption{Embedding of CASAS features.}
    \label{fig:casas_embedding}
\end{figure}

In Figure \ref{fig:casas_embedding} we show feature embeddings of the CASAS dataset. We show two diagnostic categories from opposite ends of cognitive spectrum: young volunteers and volunteers with dementia. This visualisation highlights two challenges with this dataset: 1) the class distribution is unequal (much fewer dementia data are available), and 2) there is significant overlap between the classes in this visualisation. As a result it is not surprising that difference in performance is not significant since the task is challenging. 


\begin{table}[h]
\centering
\caption{Table of results for the healthcare datasets.}
\begin{tabular}{llcc}
\toprule
Dataset & Model &         0/1 Loss &              MAE  \\
\midrule
\multirow{4}{*}{\rotatebox[origin=c]{90}{\casas{}}}
         & \bn{} &  $0.426\pm0.076$ &  $0.698\pm0.138$ \\
         & \lr{} &  $0.436\pm0.066$ &  $0.788\pm0.157$ \\
         & \ol{} &  $0.499\pm0.064$ &   $0.759\pm0.11$ \\
         & \storm{} &   $0.42\pm0.077$ &  $0.674\pm0.151$ \\
         \midrule
\multirow{4}{*}{\rotatebox[origin=c]{90}{{\sc DBank}}}
                & \bn{}  &  $0.234\pm0.045$ &   $0.382\pm0.07$ \\
                & \lr{} &  $0.228\pm0.042$ &  $0.389\pm0.075$ \\
                & \ol{} &  $0.356\pm0.038$ &  $0.456\pm0.053$ \\
                & \storm{} &  $0.234\pm0.038$ &  $0.366\pm0.063$ \\
\bottomrule
\end{tabular}
\label{table:healthcare}
\end{table}

We present the raw classification tables for the healthcare datasets in Table \ref{table:healthcare}. We can observe here that on average the 0/1 and \ac{MAE} losses are much lower on \dementiabank{} than on the \casas{} dataset. However, the losses are, on average, rather high, due to the challenging learning task.

\section{Discussion}
\label{section:discussion}

The main results presented here show that the proposed method (\storm{}) is a robust and a winning model for the prediction of ordinal quantities in most of the settings considered here. On the synthetic datasets (which primarily are used for the understanding of the model in comparison to baselines) we showed visually that our approach is able to adapt to non-linear and challenging data manifolds. Although it is highly unlikely that one will encounter manifolds of the exact form of Figure \ref{fig:synthetic} in real datasets, we also find it highly unlikely that strictly linear manifolds will be encountered in real-life scenarios. We are confident in the utility of the proposed methodology given its robust adaptation to the variation of challenging data manifolds. Although the absolute performance of all models is slightly disappointing on the healthcare datasets, this is illustrative of data representation challenges that still remain. Indeed, on these datasets some of the most important and discriminatory features (including health records) are witheald to preserve the anonymity of the participants, which further exacerbates the classification task. Yet, \storm{} is the best performing model.

\storm{} is shown to have higher performance in a statistically meaningful way on the synthetic, UCI and large datasets across all categories. In particular, we see that \storm{} achieves very good results in the large datasets (Figure \ref{fig:large}). However, in the case of the UCI datasets we see that for the 10 category dataset \storm{} is still the highest-performing model but that the the baseline ordinal regression model performs well. It is worth pointing out that many of the datasets within the UCI group were converted into an ordinal task from a regression task. Although the converted datasets still constitute legitimate ordinal challenges, we believe the process of conversion is relatively `arbitrary' and that the groupings given do not necessarily constitute meaningful groups of data. We believe this to be the reason for the absence of statistically meaningful results on the 10 category UCI datasets. However, on the large datasets we see that \storm{} is comfortably the best model amongst the baselines. We believe that this is driven primarily by the scale of the datasets here: \storm{} is better able to capture the training data distribution with larger datasets. This makes sense intuitively. Since \storm{} has a larger number of parameters these models witll typically require more data. 





\section{Conclusion}
\label{section:conclusion}

In this paper we proposed a structured propabilistic architecture for ordinal regression that is based on a structured encoding of the target variables and undirected graphical models. We have shown empirically that the proposed method (structural ordinal regression modelling) performs significantly better than three baseline methods over several synthetic, UCI and healthcare datasets. Additionally, our proposed framework has several appealing properties: inference can be vectorised over the whole dataset to speed up optimisation, locally and globally consistent abstract queries can be executed on the data, and our model preserves several desirable monotonic features for ordinal model. Future work will investigate non-linear representation methods with the proposed system and to compare the proposed techniques against more baseline methods.

\section*{Acknowledgements}
This research was conducted under the `Continuous Behavioural Biomarkers of Cognitive Impairment' project funded by the UK Medical Research Council Momentum Awards under Grant MC/PC/16029.

\bibliographystyle{unsrt}
\bibliography{main}

\balance


\section*{Appendix}

\begin{figure}
    \centering
    \subfigure[\linear{} dataset with 5 categories]{\label{fig:lin1}\includegraphics[width=0.999\linewidth]{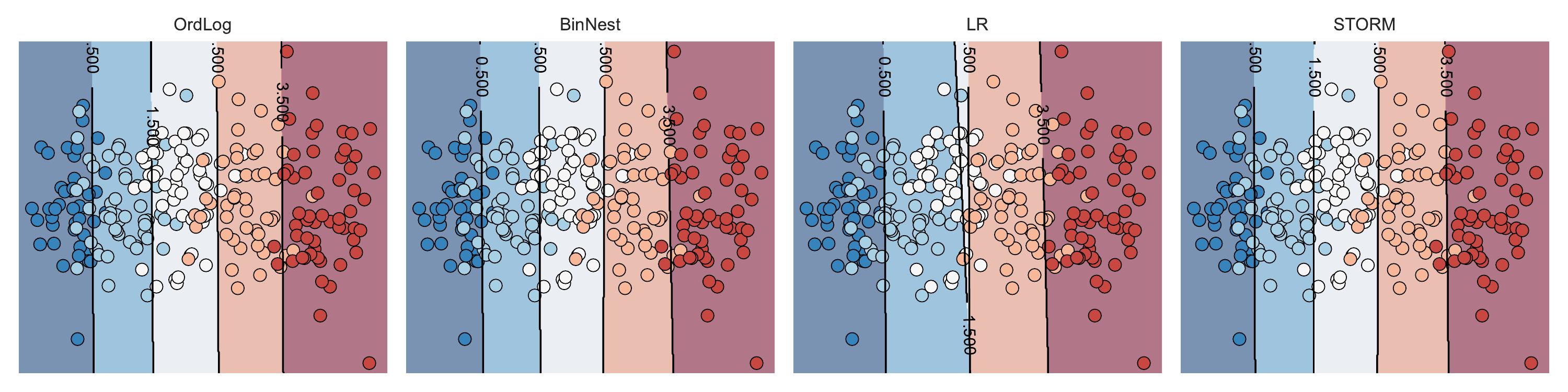}}  \\
    \subfigure[\jagged{} dataset with 5 categories]{\label{fig:lin2}\includegraphics[width=0.999\linewidth]{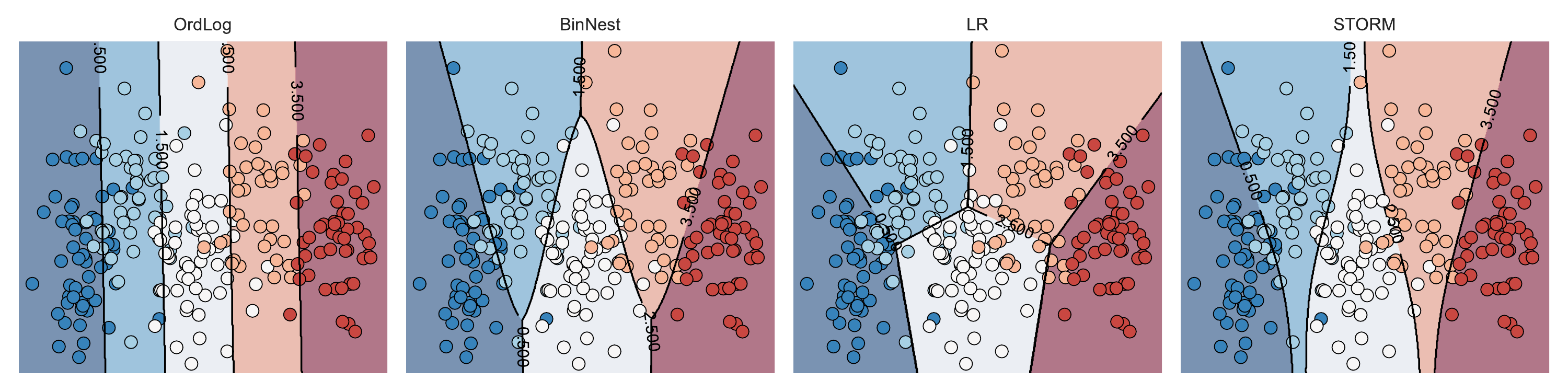}}
    \caption{
    Predictions from baseline and proposed ordinal on the \linear{} and \jagged{} datasets (Figures \ref{fig:lin1} and \ref{fig:lin2}).
    }
    \label{fig:synthetic2}
\end{figure}

Here we present additional visualisations and results tables for the interpretation and reproduction of the main results of this paper.

\subsection{Supplementary Figures}
Figure \ref{fig:synthetic2} shows the predictions of the baseline and proposed methods on the \linear{} and \jagged{} datasets.

\subsection{Supplementary Tables}
Tables \ref{table:synth5} and \ref{table:synth10} present the results on the synthetic datasets on the 5 and 10 category splits respectively, Tables \ref{table:uci5} and \ref{table:uci10} present the results on the UCI datasets on the 5 and 10 category splits respectively. The first two columns show depict the dataset and prediction model and the remining columns show the scores on 0/1 loss and MAE. 

\newpage 

\begin{table}[htb]
\centering
\caption{Table of results for the synthetic collection of datasets with 5 categories.}
\begin{tabular}{llcc}
\toprule
dataset & model &         0/1 Loss &              MAE \\
\midrule
\circle{} & \bn{} &    $0.14\pm0.01$ &  $0.174\pm0.013$\\
          & \lr{} &  $0.058\pm0.006$ &  $0.094\pm0.011$\\
          & \ol{} &  $0.519\pm0.014$ &  $0.544\pm0.014$\\
          & \storm{} &   $0.08\pm0.006$ &   $0.09\pm0.009$ \\
\jagged{} & \bn{} &  $0.128\pm0.013$ &  $0.129\pm0.013$ \\
          & \lr{} &  $0.169\pm0.013$ &  $0.173\pm0.013$ \\
          & \ol{} &  $0.164\pm0.009$ &  $0.164\pm0.009$ \\
          & \storm{} &  $0.128\pm0.011$ &  $0.128\pm0.011$ \\
\linear{} & \bn{} &    $0.17\pm0.01$ &    $0.17\pm0.01$ \\
          & \lr{} &   $0.32\pm0.018$ &  $0.322\pm0.018$ \\
          & \ol{} &   $0.17\pm0.011$ &  $0.171\pm0.011$ \\
          & \storm{} &  $0.168\pm0.011$ &  $0.168\pm0.011$\\
\spiral{} & \bn{} &  $0.299\pm0.008$ &  $0.593\pm0.011$ \\
          & \lr{} &   $0.063\pm0.01$ &  $0.066\pm0.012$ \\
          & \ol{} &   $0.65\pm0.011$ &  $0.922\pm0.016$ \\
          & \storm{} &  $0.058\pm0.009$ &    $0.06\pm0.01$ \\
\bottomrule
\end{tabular}
\label{table:synth5}
\end{table}

\begin{table}[htb]
\centering
\caption{Table of results for the synthetic collection of datasets with 10 categories.}
\begin{tabular}{llccc}
\toprule
dataset & model &         0/1 Loss &              MAE \\
\midrule
\circle{} & \bn{} &  $0.442\pm0.013$ &  $0.772\pm0.064$\\
          & \lr{} &  $0.338\pm0.012$ &  $0.641\pm0.069$\\
          & \ol{} &  $0.738\pm0.011$ &  $1.211\pm0.028$\\
          & \storm{} &   $0.356\pm0.01$ &  $0.417\pm0.021$ \\
\jagged{} & \bn{} &  $0.148\pm0.009$ &  $0.151\pm0.009$ \\
          & \lr{} &  $0.474\pm0.011$ &  $0.727\pm0.017$ \\
          & \ol{} &  $0.193\pm0.012$ &  $0.194\pm0.012$ \\
          & \storm{} &  $0.142\pm0.009$ &  $0.144\pm0.009$ \\
\linear{} & \bn{} &  $0.237\pm0.048$ &  $0.243\pm0.067$ \\
          & \lr{} &   $0.693\pm0.01$ &    $1.31\pm0.01$ \\
          & \ol{} &   $0.198\pm0.01$ &   $0.199\pm0.01$ \\
          & \storm{} &  $0.198\pm0.011$ &  $0.198\pm0.011$ \\
\spiral{} & \bn{} &    $0.68\pm0.01$ &  $2.765\pm0.043$\\
          & \lr{} &  $0.299\pm0.012$ &   $0.983\pm0.04$\\
          & \ol{} &  $0.905\pm0.007$ &  $2.474\pm0.017$\\
          & \storm{} &  $0.067\pm0.009$ &  $0.112\pm0.016$\\
\bottomrule
\end{tabular}
\label{table:synth10}
\end{table}

\begin{table}[htb]
\centering
\caption{Table of results for the UCI collection of datasets with 5 categories.}
\begin{tabular}{llcc}
\toprule
dataset & model &         0/1 Loss &              MAE \\
\midrule
\abalone{} & \bn{} &  $0.625\pm0.008$ &  $0.887\pm0.039$  \\
           & \lr{} &  $0.656\pm0.014$ &  $0.985\pm0.085$  \\
           & \ol{} &  $0.664\pm0.008$ &  $0.924\pm0.046$  \\
           & \storm{} &   $0.615\pm0.02$ &  $0.799\pm0.053$  \\
\autompg{} & \bn{} &  $0.445\pm0.028$ &  $0.549\pm0.061$ \\
           & \lr{} &  $0.498\pm0.057$ &  $0.648\pm0.173$ \\
           & \ol{} &   $0.39\pm0.019$ &  $0.394\pm0.019$ \\
           & \storm{} &  $0.361\pm0.033$ &   $0.37\pm0.036$ \\
\bostonhousing{} & \bn{} &  $0.481\pm0.076$ &  $0.618\pm0.121$ \\
           & \lr{} &  $0.564\pm0.088$ &  $0.742\pm0.149$ \\
           & \ol{} &  $0.392\pm0.048$ &   $0.48\pm0.055$ \\
           & \storm{} &  $0.337\pm0.041$ &  $0.412\pm0.052$   \\
\diabetes{} & \bn{} &   $0.739\pm0.09$ &  $0.907\pm0.135$     \\
           & \lr{} &  $0.723\pm0.058$ &  $0.949\pm0.106$      \\
           & \ol{} &  $0.643\pm0.101$ &   $0.766\pm0.12$      \\
           & \storm{} &  $0.622\pm0.129$ &  $0.771\pm0.179$   \\
\machinecup{} & \bn{} &   $0.496\pm0.08$ &  $0.657\pm0.149$   \\
           & \lr{} &  $0.553\pm0.026$ &  $0.775\pm0.113$      \\
           & \ol{} &    $0.45\pm0.08$ &  $0.509\pm0.124$      \\
           & \storm{} &  $0.414\pm0.079$ &  $0.456\pm0.087$   \\
\pyrimidines{} & \bn{} &  $0.643\pm0.081$ &  $0.791\pm0.152$  \\
           & \lr{} &  $0.608\pm0.079$ &  $0.803\pm0.143$      \\
           & \ol{} &  $0.605\pm0.087$ &   $0.73\pm0.117$      \\
           & \storm{} &   $0.503\pm0.08$ &  $0.673\pm0.154$   \\
\stocksdomain{} & \bn{} &  $0.489\pm0.098$ &  $0.603\pm0.157$ \\
           & \lr{} &   $0.39\pm0.048$ &  $0.402\pm0.052$      \\
           & \ol{} &    $0.3\pm0.023$ &  $0.304\pm0.023$      \\
           & \storm{} &  $0.146\pm0.016$ &   $0.15\pm0.016$   \\
\triazines{} & \bn{} &  $0.774\pm0.029$ &  $1.397\pm0.111$    \\
           & \lr{} &  $0.763\pm0.026$ &  $1.528\pm0.084$      \\
           & \ol{} &  $0.763\pm0.026$ &  $1.292\pm0.074$      \\
           & \storm{} &  $0.711\pm0.045$ &  $1.287\pm0.123$   \\
\wisconsin{} & \bn{} &  $0.803\pm0.032$ &  $1.535\pm0.229$    \\
           & \lr{} &  $0.797\pm0.027$ &  $1.829\pm0.223$      \\
           & \ol{} &  $0.738\pm0.051$ &  $1.159\pm0.112$      \\
           & \storm{} &   $0.813\pm0.05$ &  $1.432\pm0.154$   \\
\bottomrule
\end{tabular}
\label{table:uci5}
\end{table}

\begin{table}[htb]
\centering
\caption{Table of results for the UCI collection of datasets with 10 categories.}
\begin{tabular}{llccc}
\toprule
dataset & model &         0/1 Loss &              MAE \\
\midrule
\abalone{} & \bn{} &  $0.831\pm0.025$ &  $2.389\pm0.126$ \\
           & \lr{} &  $0.808\pm0.004$ &  $2.306\pm0.122$ \\
           & \ol{} &  $0.796\pm0.004$ &  $1.671\pm0.122$ \\
           & \storm{} &  $0.747\pm0.015$ &  $1.572\pm0.133$ \\
\autompg{} & \bn{} &  $0.728\pm0.043$ &  $1.342\pm0.244$    \\
           & \lr{} &   $0.786\pm0.05$ &  $2.268\pm0.582$    \\
           & \ol{} &   $0.57\pm0.044$ &   $0.85\pm0.091$    \\
           & \storm{} &  $0.544\pm0.046$ &  $0.753\pm0.115$ \\
\bostonhousing{} & \bn{} &   $0.719\pm0.02$ &  $1.426\pm0.041$\\
           & \lr{} &  $0.786\pm0.035$ &  $1.999\pm0.245$    \\
           & \ol{} &  $0.577\pm0.031$ &  $0.858\pm0.081$    \\
           & \storm{} &  $0.559\pm0.054$ &   $0.84\pm0.109$ \\
\diabetes{} & \bn{} &  $0.808\pm0.086$ &  $1.712\pm0.417$\\
           & \lr{} &   $0.844\pm0.03$ &   $1.731\pm0.18$ \\
           & \ol{} &  $0.838\pm0.084$ &  $1.473\pm0.278$ \\
           & \storm{} &   $0.787\pm0.14$ &  $1.619\pm0.317$ \\
\machinecup{} & \bn{} &  $0.708\pm0.088$ &   $1.699\pm0.69$ \\
           & \lr{} &  $0.792\pm0.046$ &  $2.802\pm0.533$ \\
           & \ol{} &  $0.569\pm0.094$ &   $0.984\pm0.31$ \\
           & \storm{} &  $0.562\pm0.084$ &  $0.989\pm0.269$  \\
\pyrimidines{} & \bn{} &  $0.715\pm0.093$ &  $1.169\pm0.302$ \\
           & \lr{} &  $0.735\pm0.092$ &  $1.763\pm0.438$ \\
           & \ol{} &  $0.658\pm0.083$ &  $1.045\pm0.162$ \\
           & \storm{} &  $0.613\pm0.091$ &  $1.086\pm0.313$   \\
\stocksdomain{} & \bn{} &  $0.639\pm0.051$ &  $1.059\pm0.229$ \\
           & \lr{} &   $0.813\pm0.07$ &  $2.048\pm0.389$ \\
           & \ol{} &  $0.579\pm0.022$ &  $0.657\pm0.023$ \\
           & \storm{} &  $0.296\pm0.019$ &  $0.308\pm0.021$ \\
\triazines{} & \bn{} &   $0.853\pm0.02$ &  $2.647\pm0.199$  \\
           & \lr{} &  $0.885\pm0.013$ &  $2.929\pm0.206$ \\
           & \ol{} &   $0.85\pm0.028$ &  $2.301\pm0.235$ \\
           & \storm{} &  $0.816\pm0.047$ &  $2.371\pm0.367$ \\
\wisconsin{} & \bn{} &  $0.901\pm0.033$ &   $3.292\pm0.55$ \\
           & \lr{} &  $0.898\pm0.011$ &  $4.268\pm0.293$\\
           & \ol{} &  $0.874\pm0.035$ &  $2.469\pm0.236$\\
           & \storm{} &  $0.918\pm0.039$ &   $3.051\pm0.28$\\
\bottomrule
\end{tabular}
\label{table:uci10}
\end{table}


\end{document}